\documentclass{article}

\usepackage{microtype}
\usepackage{graphicx}
\usepackage{subcaption}
\usepackage{booktabs} 

\usepackage{hyperref}

\usepackage[preprint]{icml2026}

\usepackage{amsmath}
\usepackage{amssymb}
\usepackage{mathtools}
\usepackage{amsthm}

\usepackage[capitalize,noabbrev]{cleveref}

\usepackage{enumitem}

\theoremstyle{plain}

\theoremstyle{definition}

\theoremstyle{remark}

\usepackage[textsize=tiny]{todonotes}

\icmltitlerunning{Emergent Search and Backtracking in Latent Reasoning Models}

\begin{document}

\twocolumn[
  \icmltitle{Emergent Search and Backtracking in Latent Reasoning Models}


  \icmlsetsymbol{equal}{*}

  \begin{icmlauthorlist}
    \icmlauthor{Jasmine Cui}{equal,ind}
    \icmlauthor{Charles Ye}{equal,ind}
  \end{icmlauthorlist}

  \icmlaffiliation{ind}{Independent}

  \icmlcorrespondingauthor{Jasmine Cui}{dogdynamics@proton.me}

  \icmlkeywords{Machine Learning, Latent Reasoning, Adaptive Compute, Latent Reasoning Models, Universal Transformers}

  \vskip 0.3in
]



\printAffiliationsAndNotice{}  

\begin{abstract}
What happens when a language model thinks without words? Standard reasoning LLMs verbalize intermediate steps as chain-of-thought; \textbf{latent reasoning transformers} (LRTs) instead perform deliberation entirely in continuous hidden space.

We investigate an LRT, decoding the model's evolving beliefs at every step on a multiple-choice QA benchmark. We find that the model spontaneously learns \textbf{a structured search process in latent space}. Deliberation follows a consistent trajectory: an exploration phase where probability mass spreads across candidates, tentative commitment to a frontrunner, and either convergence or \textbf{backtracking}. Backtracking is prevalent (32\% of instances), beneficial (34\% accuracy gain over non-backtracking instances), and predominantly directed away from the semantically closest distractor toward the correct answer. The search is adaptive: replacing distractors with implausible alternatives shortens exploration by 54\%. Latent reasoning models achieve in activation space what chain-of-thought achieves through words: the ability to be wrong, notice, and recover.
\end{abstract}


\section{Introduction}
\label{sec:intro}
Reasoning LLMs solve hard problems by thinking step by step — generating intermediate text before committing to an answer. This is \textbf{adaptive compute}: the model allocates additional computation by using each step's output to inform the next. But generating in natural language forces deliberation through a bottleneck, constraining every intermediate thought through language.

\textbf{Latent reasoning transformers} (LRTs) avoid this bottleneck by thinking without words. LRTs implement adaptive computation entirely in latent space, forgoing generation of intermediate text altogether. Research in LRTs has been motivated by the efficiency benefits of avoiding natural language generation, and the expressivity benefits of latent thought for tasks that language cannot easily serialize.

Yet precisely because LRTs think without words, we know remarkably little about how they think. Do they converge linearly to an answer? Do they explore alternatives? Can they change their minds?

\begin{figure}[t]
\begin{center}
\centerline{\includegraphics[width=\columnwidth]{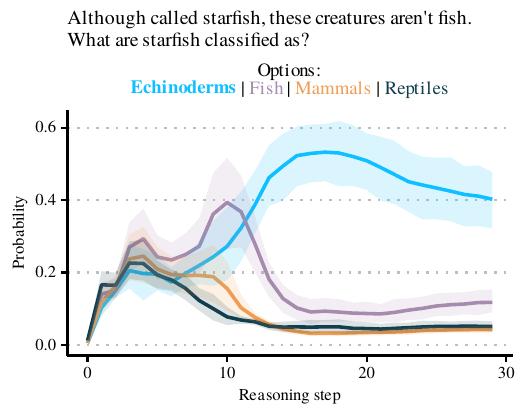}}
\caption{\textbf{Latent thought on a single question.} The model explores candidates roughly uniformly, commits to the semantically closest distractor (\textit{fish}), then backtracks to the correct answer (\textit{echinoderm}). This trajectory — exploration, shallow commitment, error correction — is representative of the search dynamics we characterize across the benchmark. Shaded regions = 95\% CIs over 25 random answer-order permutations.}
\vspace{-1em}
\label{fig:fig1}
\end{center}
\end{figure}

A key property makes these answerable. Looped LRTs iterate by passing the hidden state through the same transformer block repeatedly \citep{dehghani2019universaltransformers, chen2025innerthinkingtransformerleveraging, zhu2025surveylatentreasoning, zhu2025scalinglatentreasoninglooped, geiping2025scalingtesttimecomputelatent}. Because they are trained at variable recurrence depths, every intermediate state is decoder-ready — not approximately, but exactly. We can decode the model's token predictions at every step of deliberation, yielding a frame-by-frame recording of evolving beliefs, as seen in \Cref{fig:fig1}.

\begin{figure*}[t]
\begin{center}
\centerline{\includegraphics[width=0.95\linewidth]{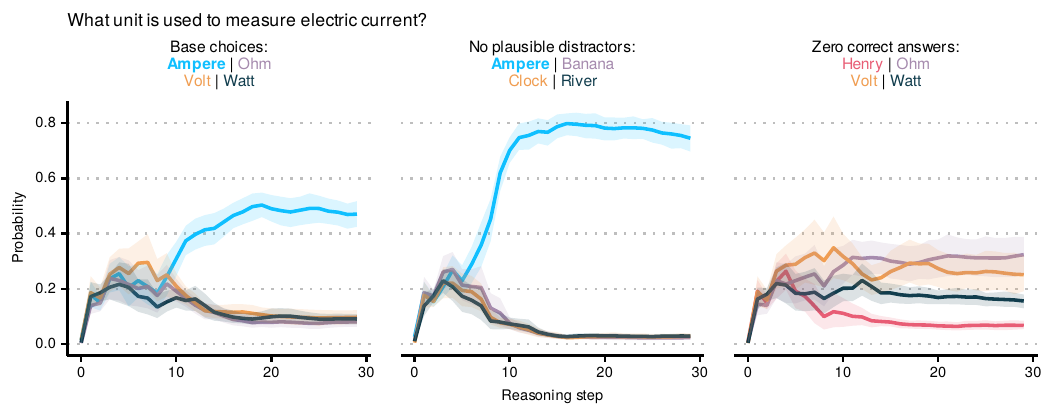}}
\caption{\textbf{Task difficulty modulates deliberation.}
Belief trajectories for the same question under three
answer-set variants. \textit{Base} (plausible distractors):
the model explores before gradually converging.
\textit{Easy} (unrelated distractors): convergence is fast
and confident. \textit{No correct answer}: probability mass
remains distributed --- the model never commits. Shaded
regions = 95\% CIs over 25 answer-order permutations.}
\vspace{-1em}
\label{fig:variants}
\end{center}
\end{figure*}

Our findings:
\begin{itemize}
    \item Latent reasoning follows a \textbf{structured search process}: an exploration phase where probability mass spreads across candidates, tentative commitment to a frontrunner, and either convergence or backtracking. The search is adaptive — causally reducing task difficulty shortens exploration by 54\%.
    \item Spontaneous \textbf{backtracking} occurs in 32\% of instances, improves accuracy by 34\%, and is systematic — the abandoned answer is predominantly the semantically closest distractor to the question.
\end{itemize}

Latent thought gives the model what chain-of-thought provides through words: the ability to be wrong, notice, and recover.

\paragraph{Model.}
We study \textsc{Huginn-0125} \citep{geiping2025scalingtesttimecomputelatent}, a 3.5B-parameter looped LRT. The architecture consists of a \textbf{prelude} $P$ (embedding layer and 2 transformer layers), a \textbf{recurrent block} $R$ (4 transformer layers, iterated $K$ times), and a \textbf{coda} $C$ (2 transformer layers and the LM head):
\begin{equation}
  h_0 = P(x), \qquad h_i = R(h_{i-1}) \;\text{for}\; i = 1, \ldots, K.
\end{equation}
Because the model is trained with randomly sampled recurrence depths, the coda has consumed states $h_i$ for every $i$ during training. We decode the model's predictive distribution at every step by passing each intermediate state through the coda:
\begin{equation}
  p_i = \mathrm{softmax}\!\big(C(h_i)\big), \qquad i = 1, \ldots, K.
\end{equation}
Each $p_i$ is the distribution the model would genuinely output
if halted at step $i$ --- not a heuristic projection, but an exact
readout. We track the probability assigned to each of the four
answer options across steps, yielding the belief trajectories
shown in \Cref{fig:fig1}.

\paragraph{Dataset.}
We construct a 260-item four-choice QA benchmark spanning factual recall, definitions, multi-step logic, arithmetic, and adversarially misleading questions. Each question stem is paired with three answer-set variants:
\begin{itemize}[nosep]
  \item \textbf{Base}: all distractors are plausible and
        semantically related to the question.
  \item \textbf{Easy}: incorrect options are obviously unrelated
        to the question stem.
  \item \textbf{No correct answer}: the correct option is replaced
        with an additional distractor.
\end{itemize}
These variants are causal manipulations of task difficulty: comparing Base to Easy isolates the effect of distractor plausibility on deliberation dynamics, while No correct answer tests whether the model converges when no valid option exists. Each question--variant pair is evaluated under 25 random permutations of the answer order
to control for position bias. We run $K = 30$ recurrent steps per permutation, decoding $p_i$ at every step. All results average over permutations; shaded regions in figures show 95\% bootstrap CIs.

\section{The Model Searches}
\label{sec:results}

\subsection{Phased Deliberation}
\label{sec:phases}

Decoding $p_i$ at every step reveals that the model does not
converge linearly to an answer. Instead, deliberation follows a
consistent multi-phase trajectory (\Cref{fig:variants}):

\begin{enumerate}[nosep]
  \item \textbf{Exploration.} Probability mass is spread roughly
        uniformly across options. The model has not yet
        differentiated candidates.
  \item \textbf{Shallow commitment.} One option begins to dominate,
        typically the distractor most semantically related to the
        question stem --- a surface-level match rather than the
        correct answer.
  \item \textbf{Convergence.} The model either strengthens its
        commitment to the leading option, or \emph{backtracks}:
        abandoning the frontrunner and shifting to a different
        answer (Section~\ref{sec:backtracking}).
\end{enumerate}

\paragraph{The search is difficulty-adaptive.}
If the phased structure reflects genuine deliberation, harder
problems should require more exploration. We test this by comparing
the Base and Easy variants, which share the same question stem and
correct answer but differ only in distractor plausibility.

\begin{figure}[t]
\begin{center}
\centerline{\includegraphics[width=\columnwidth]{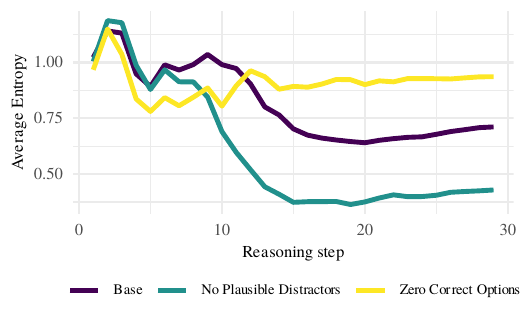}}
\caption{\textbf{Entropy tracks task difficulty.}
Average entropy of $p_i$ across recurrence steps, by variant. Easy questions converge rapidly to low entropy. Base questions converge more slowly. No correct answer questions remain at high entropy --- the model recognizes persistent uncertainty.}
\label{fig:entropy}
\end{center}
\end{figure}

We operationalize the exploration phase as the initial window
during which the model's beliefs are actively changing. Specifically,
we measure the KL divergence between consecutive decoded
distributions:
\begin{equation}
  D_{\mathrm{KL}}(p_{i+1} \| p_i)
    = \sum_{v \in \mathcal{V}} p_{i+1}(v) \log \frac{p_{i+1}(v)}{p_i(v)},
\end{equation}
and define exploration as ending when
$D_{\mathrm{KL}}(p_{i+1} \| p_i) \leq 0.01$ for 3 consecutive steps.
By this criterion, Base questions explore for an average of
54\% more steps than their Easy counterparts
(\Cref{fig:entropy}). The model allocates more deliberation
to problems that demand it.

The No correct answer variant provides a complementary test:
when no valid option exists, the model should never reach stable
commitment. Indeed, entropy remains elevated throughout all 30
steps, and the trajectory never enters a low-entropy attractor
(\Cref{fig:entropy}). The model's deliberation is not
a fixed-length warmup --- it is a search that terminates when
(and only when) it finds something.

\subsection{Backtracking}
\label{sec:backtracking}

The most striking emergent behavior is \emph{backtracking}: the
model commits to one answer, then reverses itself. We define a
backtracking event as a trajectory in which the argmax answer
is $a$ for at least 3 consecutive steps, later becomes $b \neq a$
for at least 3 consecutive steps, and the final answer is $b$.

Backtracking is prevalent, beneficial, and systematic.

\paragraph{Prevalent.} 32\% of Base-variant instances exhibit
at least one backtracking event.

\paragraph{Beneficial.} Instances that backtrack achieve 34\%
higher accuracy than non-backtracking instances. The model is
more accurate specifically because it revises.

\paragraph{Systematic.} What does the model backtrack \emph{from}?
If backtracking reflects genuine error correction --- rejecting a
shallow match in favor of a deeper one --- the abandoned answer
should disproportionately be the distractor most semantically
similar to the question stem. We test this by computing the
cosine similarity between the sentence embedding of the question
stem and each answer option, then checking whether the abandoned
answer in backtracking events is the highest-similarity distractor.

\begin{table}[t]
\caption{\textbf{Backtracking is directed.} In backtracking events,
the abandoned answer is the semantically closest distractor far
more often than chance (25\% for four options). The model
corrects away from shallow semantic matches toward the correct
answer.}
\label{tab:backtrack}
\begin{center}
\begin{tabular}{lc}
\toprule
\textbf{Abandoned answer is\ldots} & \textbf{\% of backtracks} \\
\midrule
Most similar distractor  & 72\% \\
2nd most similar         & 22\% \\
Least similar distractor & 6\% \\
\midrule
Adopted answer is correct & 52\% \\
\bottomrule
\end{tabular}
\end{center}
\end{table}

\Cref{tab:backtrack} confirms the pattern: the abandoned answer
is the most semantically similar distractor in 72\% of
backtracking events, far above the 25\% chance baseline. In
52\% of cases, the model backtracks \emph{to} the correct answer.
Backtracking is not random exploration --- it is directed error
correction, moving from surface similarity to the right answer.
\section{Discussion}
\label{sec:discussion}

\paragraph{What iteration buys.}
Thinking structure emerges spontaneously: cheap shallow filtering in early steps, refinement in later steps, and error correction when the shallow filter fails.

The 72\% figure makes the mechanism concrete. The model's initial commitment is dominated by surface-level semantic matching --- it gravitates toward the distractor closest to the question in embedding space. Its later steps override this with a deeper assessment, rejecting the shallow match in favor of the correct answer. Iterative computation enables a two-timescale strategy: fast approximate retrieval, followed
by slower verification and correction. Backtracking is the observable signature of this correction.

\paragraph{Scope.}
We study a single 3.5B-parameter model on a 260-item synthetic benchmark with four-choice questions. Whether the same dynamics appear in larger LRTs, on open-ended generation tasks, or in non-looped adaptive-compute architectures is untested.

\paragraph{Reframing chain-of-thought.}
Chain-of-thought also provides iterative computation — each generated token feeds back into the next step. If the underlying computational need is search, CoT models may
implement similar dynamics: exploring candidates, committing tentatively, and revising. But CoT entangles this search with a second objective — generating human-legible text — and the faithfulness literature suggests these come apart \citep{turpin2023unfaithful, zhou2024causal}. Our results offer a complementary perspective: when the language
bottleneck is removed, the latent computation that remains is structured search. This suggests that unfaithful chain-of-thought may be a lossy narration of a search process that is, at its source, structured.

\paragraph{Interpretability of LRTs.} Because every recurrent state is decoder-ready, these dynamics are not inferred or approximated — they are directly observed in the model's own predictive space. This makes looped LRTs a uniquely transparent form of adaptive compute: the reasoning is the trajectory, and the trajectory is readable.

\paragraph{Conclusion.}
We decoded every step of a latent reasoning model's deliberation and found that it searches: exploring candidates, committing on the basis of shallow similarity, and spontaneously reversing when that commitment is wrong. This search is not prescribed by
the architecture or the training objective — it is discovered by the model as an efficient use of iterative computation.

Latent thought gives the model what chain-of-thought provides through words: the ability to be wrong, notice, and recover.
\vfill
\bibliography{main}
\bibliographystyle{icml2026}

\newpage
\appendix
\onecolumn

\clearpage


\end{document}